# Model-Informed Generative Adversarial Network (MI-GAN) for Learning Optimal Power Flow

Yuxuan Li[1], Chaoyue Zhao[2], and Chenang Liu[1*]

[1]The School of Industrial Engineering & Management, Oklahoma State University, Stillwater, OK

[2]Department of Industrial and Systems Engineering, University of Washington, Seattle, WA

*Corresponding author: chenang.liu@okstate.edu

**Abstract:** The optimal power flow (OPF) problem, as a critical component of power system operations, becomes increasingly difficult to solve due to the variability, intermittency, and unpredictability of renewable energy brought to the power system. Although traditional optimization techniques, such as stochastic and robust optimization approaches, could be leveraged to address the OPF problem, in the face of renewable energy uncertainty, i.e., the dynamic coefficients in the optimization model, their effectiveness in dealing with large-scale problems remains limited. As a result, deep learning techniques, such as neural networks, have recently been developed to improve computational efficiency in solving OPF problems with the utilization of data. However, the feasibility and optimality of the solution may not be guaranteed, and the system dynamics cannot be properly addressed as well. In this paper, we propose an optimization model-informed generative adversarial network (MI-GAN) framework to solve OPF under uncertainty. The main contributions are summarized into three aspects: (1) to ensure feasibility and improve optimality of generated solutions, three important layers are proposed: feasibility filter layer, comparison layer, and gradient-guided layer; (2) in the GAN-based framework, an efficient model-informed selector incorporating these three new layers is established; and (3) a new recursive iteration algorithm is also proposed to improve solution optimality and handle the system dynamics. The numerical results on IEEE test systems show that the proposed method is very effective and promising.

# 1 Introduction

The national electricity sector has witnessed a dramatic change in renewable energy penetration in recent years, and its share is expected to continue growing rapidly in the next few decades. On a federal level, the U.S. Energy Information Administration (EIA) predicts that renewable energy resources will provide 38% of electricity energy by 2050, with solar and wind energy accounting for 17.5% and 12.54%, respectively (Outlook 2009). With such a high renewable energy penetration, it is critical and urgent for power system operators to learn how to operate power systems effectively and securely in the face of high variability, intermittency, and unpredictability of renewable energy output.

Optimal power flow (OPF), as an essential component of power system operations and management, aims to minimize the total generation cost while satisfying a series of system constraints and operational requirements, achieving economic power system operation in day-ahead and real-time markets. As a key building block of many complex power system problems, the optimization approaches for solving OPF problems has been extensively studied in recent years. The most common approach to formulating the OPF problem with renewable energy uncertainties is stochastic OPF (e.g., (Yong and Lasseter 2000; Kimball et al. 2003) among others), in which the uncertain renewable energy output is characterized by a finite number of scenarios, or is assumed to follow a particular probability distribution, and then sampling approaches are typically used to finalize the formulation. The robust optimization approach (e.g., (Delage and Ye 2010; Martinez-Mares and Fuerte-Esquivel 2013) among others) is another traditional approach to addressing the OPF problem with renewable energy uncertainties. Its key idea is to construct an OPF solution that is optimal for the worst-case realization of renewable energy output in a predefined uncertainty set. Recently, the distributionally robust OPF problem (e.g., (Zhang, Shen, and Mathieu 2016; Yi Guo et al. 2018) among others) has also been proposed for hybrid stochastic and robust approaches, by considering the worst-case distribution of renewable energy output. The obtained OPF decisions are expected to be less conservative than the ones from the robust approach and more reliable than the ones from the stochastic approach.

Despite the great success of these approaches in effectively addressing OPF problem with the renewable energy uncertainty, the practical use of these methods remains limited due to scalability issues for large-scale power systems. In recent years, the staggering advances obtained in deep learning make it possible to quickly find the optimal strategy of computationally intensive power system optimization problems with the utilization of data. Over the last decade, various deep learning approaches have been used to address OPF problems and can be divided into two major categories. One is based on the training of neural networks, such as deep neural networks (DNN) (Nellikkath and Chatzivasileiadis 2022; Fioretto, Mak, and Van Hentenryck 2020; Yang, Gao, and Fan 2021), convolutional neural networks (CNN) (Jia and Bai 2021; Zhao et al. 2023) and graph neural networks (GNN) (Falconer and Mones 2022; Owerko, Gama, and Ribeiro 2020; S. Liu, Wu, and Zhu 2022), to learn a high-dimensional mapping between net load (i.e., load minus renewable energy output) inputs and corresponding dispatch and transmission decisions from historical records. However, they primarily depend on supervised learning techniques and massive simulations obtained ahead of time to train the neural networks. As for the approaches that could acquire the solutions fast (Donti, Rolnick, and Kolter 2021; Chatzos, Mak, and Van Hentenryck 2021), they still need to be re-trained with updated feasible solutions as the net loads vary in the power systems. The other direction is to utilize the deep reinforcement learning (RL) to generate instantaneous OPF decisions corresponding to the real-time change of net load inputs (Yan and Xu 2020; Zhou et al. 2020). However, since RL algorithms are known as learning by trial and error, the obtained OPF controls also lack a safety guarantee.

To address the OPF problems, we propose a novel optimization model-informed generative adversarial network (MI-GAN) framework in this paper. The emerging generative adversarial network (GAN) was first proposed by Goodfellow (Goodfellow et al. 2014). It was originally designed for image augmentation (Tanaka and Aranha 2019), but it is now applied to a variety of fields, including manufacturing (Li et al. 2021), business fraud detection (Fiore et al. 2019), and healthcare (Y. Liu et al. 2019). GAN trains two networks, generator $G$ and discriminator $D$, based on a minimax game for $V(D, G)$ shown in Eq. (1).

$$\min_{G} \max_{D} V(D, G) = \mathbb{E}_{\mathbf{x} \sim P_{\text{data}}(\mathbf{x})}[\log(D(\mathbf{x}))] + \mathbb{E}_{\mathbf{z} \sim P_Z(\mathbf{z})}[\log(1 - D(G(\mathbf{z})))] \qquad (1)$$

where $G$ generates artificial samples $G(\mathbf{z})$ based on noise $\mathbf{z}$ generated at random. Then $D$ differentiates between artificial samples $G(\mathbf{z})$ and actual samples $\mathbf{x}$. Following that, $G$ and $D$ compete against each other. The ultimate goal of GAN is to make it nearly impossible for $D$ to tell whether the generated samples are actual data or artificial data. Then the artificial data is sufficiently similar to the actual data (Goodfellow et al. 2014). In this way, even when the data is complex and high-dimensional, a GAN-based framework can learn the underlying relationships among the data. Deep convolutional GAN (DCGAN) (Radford, Metz, and Chintala 2016), for example, could generate high-dimensional images with competitive performance by incorporating a convolutional neural network (CNN). Furthermore, conditional GAN (CGAN) (Douzas and Bacao 2018) makes the training process more stable and controllable by using target classification variables as a reference in iterations.

GAN-based approaches, in particular, could also be used in optimization. For instance, Tan *et al.* (Tan and Shi 2019) proposed Generative Adversarial Optimization (GAO) based on the GAN model, which is also applied in many applications, including acute lymphocytic leukemia detection (Tuba and Tuba 2019), and crowdsensing (Yi-nan Guo et al. 2020). It considers the output of generator as generated solutions while the discriminator estimates the probability that the generated solution is better than the current solution. However, the performance of GAO may be limited when the optimization problem is complex due to its functionality by reducing the radius of solution searching space. In addition, Wang *et al.* (Wang and Srikantha 2022) has also proposed a novel generative framework for OPF problems. It utilizes generator to generate feasible solutions and verify the feasibility of solutions by discriminator. However, it requires the autoencoder to handle the optimality. Therefore, a natural idea to apply the GAN-based framework for handling both feasibility and optimality is demonstrated as follow: The output of generator $G$ could be generated solutions, whereas the actual data could be the historical solutions, i.e., the actual solutions centered on the optimal solution. As a result, the distribution of generated solutions could gradually approach to the distribution of historical solutions. Then the near-optimal solution could be obtained.

However, if the historical solutions do not exist or are not centered on optimal solutions, updating the generated solutions to move forward to the optimal solution becomes a challenge. Besides, ensuring the generated solutions by $G$ to satisfy the constraints is difficult.

In this paper, inspired by our previous work of augmented time-regularized GAN (ATR-GAN) (Li et al. 2021), we leverage the proposed model-informed GAN (MI-GAN) framework to address the OPF problems with uncertainties. Specifically, the uncertainties in the power system are demonstrated from the dynamic changes of the net loads in the direct current OPF (DC-OPF) problem. The proposed novel MI-GAN differs from traditional deep learning approaches in two major ways.

First, rather than learning the mapping between the net load and control decisions, the GAN architecture can learn the underlying distribution of optimal OPF decisions and directly generate corresponding solutions for different realizations of net load. In this manner, solutions can be fast generated from the learned distribution without involving the neural network. Second, unlike traditional neural network-based methods, including the conventional GAN, which ignore the model structure, the proposed MI-GAN incorporates the model constraints and gradient information during the training of GAN in the following ways: (i) to ensure the feasibility, a new feasibility filter layer is added to the model-informed selector to check if the solution meets all the physical constraints. If not, the solution will be filtered out, and this information will be synthesized to MI-GAN to discourage similar solutions from being generated again; (ii) to utilize the information from the objective function and further improve the solution optimality, the resultant objective function value and gradient information are also integrated into the generator training to guide MI-GAN to search for better solutions; and (iii) MI-GAN does not require a large number of data samples, since it can guide the search and improve its solution quality using self-generated data.

To summarize, the contributions of the proposed method consist of three parts:

1) To address OPF problems with dynamic net loads, three critical new layers are designed to ensure the feasibility and improve the optimality of generated solutions: the feasibility filter layer, the comparison layer, and the gradient-guided layer.

2) A new GAN-based learning framework, MI-GAN, is proposed by incorporating an efficient model-informed selector, which is designed to integrate the three proposed novel layers.

3) A new recursive iteration algorithm is also proposed to further reduce the bias between selected solutions and optimal solutions.

Furthermore, the performance of the proposed MI-GAN framework is investigated using multiple testing cases, and the results show the supreme performance and efficiency of our proposed method compared with optimization and the state-of-art AI-optimization approaches on OPF problems. The rest of the work is organized as follows: In Section 2, the general mathematical formulation of the DC-OPF problem is presented. In Section 3, model-informed GAN is proposed to solve the DC-OPF problems. Numerical results and comparisons on IEEE test systems are presented in Section 4 to demonstrate the effectiveness of the proposed method. Finally, conclusions are drawn in Section 5.

## 2 The Optimal Power Flow Problem

In this study, the direct current OPF (DC-OPF) problem with dynamic power load is applied for the purpose of concept proof. In the following DC-OPF problem, we assume that the voltages are constants at all buses. Then there are two groups of variables: the power generation, $\mathbf{p}_g$, and the power voltage phase angles, $\boldsymbol{\theta}$. Denote $n_b$ and $n_g$ as the numbers of buses and power generators respectively. Then $\mathbf{p}_g = \{p_{g1}, p_{g2}, \ldots, p_{gn_g}\}$ where $p_{gi}$ is the power generation for generator $i$, and $\boldsymbol{\theta} = \{\theta_1, \theta_2, \ldots, \theta_{n_b}\}$ where $\theta_i$ is the phase angle for bus $i$. In addition, denote that the power load is $\mathbf{p}_d$ and there are $n_d$ buses that have the net load.

Specifically, to quantify the uncertainties in the DC-OPF model, the power load $\mathbf{p}_d$ has a coefficient $\rho$. $\rho$ is considered as 1 when starting to solve this DC-OPF model. Afterwards, $\rho$ may also change to other values in $(0, +\infty)$ so that the net loads may change due to the renewable energy uncertainty. Thus, the level of uncertainty can be quantified by the scale of $\rho$. Under such circumstances, the OPF problem may contain a series of optimization models to be solved where $\rho$ corresponds to different values.

Besides, based on the lines between different buses, the power output could be sent following one matrix

$\mathbf{M}_g$ and the power load could be received following another matrix $\mathbf{M}_d$. The size of $\mathbf{M}_g$ is $n_b \times n_g$ while the size of $\mathbf{M}_d$ is $n_b \times n_d$. Then, the formulation of the DC-OPF problem can be expressed by Eq. (2)-(5).

$$\min_{\mathbf{p}_g, \boldsymbol{\theta}} \mathbf{c}^T \mathbf{p}_g \tag{2}$$

$$\text{s.t. } \mathbf{M}_g \mathbf{p}_g - \mathbf{M}_d(\rho \mathbf{p}_d) = \mathbf{B}_{\text{bus}} \boldsymbol{\theta} \tag{3}$$

$$\boldsymbol{p}_{\text{line}}^{\min} \leq \mathbf{B}_{\text{line}} \boldsymbol{\theta} \leq \boldsymbol{p}_{\text{line}}^{\max} \tag{4}$$

$$\boldsymbol{p}_g^{\min} \leq \mathbf{p}_g \leq \boldsymbol{p}_g^{\max} \tag{5}$$

Eq. (2) is the function to minimize the power generation cost where the coefficients used to calculate the cost for all the generators are denoted as $\mathbf{c}$. Following that, the minimization is constrained by the power balance equation, the transmission line capacity constraints, and the generation operation limits. Eq. (3) means that the power output should balance the net load, and the in- and out-going flows for all the buses where $\mathbf{B}_{\text{bus}}$ is the bus admittance matrix. In addition, Eq. (4) shows the minimum and maximum active line flow limits based on the line admittance matrix $\mathbf{B}_{\text{line}}$. Eq. (5) demonstrates the minimum and maximum power generation limits.

To simplify the expressions, the solution is denoted as $\mathbf{x}$ and the objective function is denoted as $f(\cdot)$. Since the DC-OPF problem is a linear programming problem, the constraints are represented by $\mathbf{Ax} \leq \mathbf{b}$ in Sec. 3. Notably, the constraints for AC-OPF problems could also be addressed by the proposed MI-GAN if considering the constraints as $g(\mathbf{x}) \leq \mathbf{b}$ instead.

## 3 Research Methodology

### 3.1 Model-informed GAN (MI-GAN)

As illustrated in Figure 1, the MI-GAN is proposed to involve the OPF model in the training of GAN. It includes two key components, a model-informed (MI) generator $G_m$ and a discriminator $D$. Based on the adversarial architecture in GAN, $G_m$ will first generate the OPF solutions, and $D$ will identify if the input solutions are generated or actual. $G_m$ and $D$ will compete until the distribution of generated solutions resembles the distribution of actual solutions. In this framework, the historical OPF solutions, $\mathbf{X}_h$, i.e., the optimal or near optimal solutions of the previous/solved OPF models, are considered as the actual solutions to feed $D$. Besides, a preserved matrix is also applied to save the generated solutions from the proposed MI-GAN, and this matrix is defined as saved solutions, i.e., $\mathbf{X}_s$.

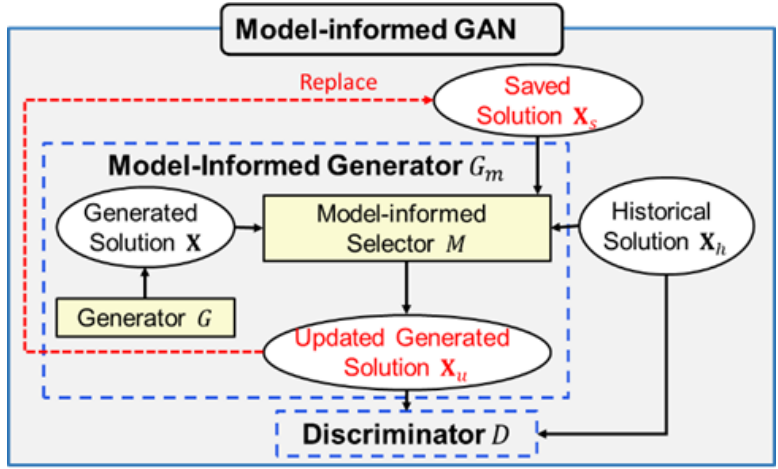

Figure 1 The overview of the proposed MI-GAN.

According to Eq. (1) in Sec. 1, the minimax game for the MI-GAN, i.e., $V(D, G_m)$, is formulated in Eq. (6),

$$\min_{G_m} \max_{D} V(D, G_m) = \mathbb{E}_{\mathbf{x}_h \sim P_{\mathbf{data}}(\mathbf{x}_h)}[\log(D(\mathbf{x}_h))] + \mathbb{E}_{\mathbf{z} \sim P_{\mathbf{Z}}(\mathbf{z})}[\log(1 - D(G_m(\mathbf{z})))] \qquad (6)$$

where $\mathbf{x}_h$ is the historical solution while $\mathbf{z}$ is the input noise for $G_m$. Accordingly, the loss of $G_m$, i.e., $L_{G_m}$, and the loss of $D$, $L_D$, could be demonstrated in Eq. (7).

$$L_{G_m} = -\mathbb{E}_{\mathbf{z} \sim P_{\mathbf{Z}}(\mathbf{z})}(D(G_m(\mathbf{z})))$$
$$L_D = \mathbb{E}_{\mathbf{z} \sim P_{\mathbf{Z}}(\mathbf{z})}(D(G_m(\mathbf{z}))) - \mathbb{E}_{\mathbf{x}_h \sim P_{\mathbf{data}}(\mathbf{x}_h)}(D(\mathbf{x}_h)) \qquad (7)$$

$G_m$ consists of a regular generator $G$ and a newmodel-informed (MI) selector $M$. As shown in Eq. (8) in each training iteration, $G$ will first generate solutions, i.e., $G(\mathbf{z})$, and then it will be sent to $M$. Afterwards, according to the OPF constraints, $M$ will select and update the feasible solutions among the generated solutions, based on the saved solution, $\mathbf{X}_s$, and historical solution $\mathbf{X}_h$.

$$G_m(\mathbf{z}) = M(G(\mathbf{z})|\mathbf{X}_s, \mathbf{X}_h) \qquad (8)$$

Subsequently, both $\mathbf{X}_h$ and the selected generated solutions i.e., $\mathbf{X}_u$, will be sent to $D$. Based on the output of $D$, the losses $L_{G_m}$ and $L_D$ could be obtained to update the model during iteration. $\mathbf{X}_u$ will be saved and applied as $\mathbf{X}_s$ in next iteration. In this way, $\mathbf{X}_s$ and $(D, G_m)$ can be updated iteratively until it converges.

Specifically, the input of the regular generator $G$ is randomly generated noise $\mathbf{z}$ and the output is the solution $G(\mathbf{z})$. Besides, the input of the discriminator $D$ is $G(\mathbf{z})$ and $\mathbf{x}_h$, while the output is a vector to measure the similarity between actual and generated solutions. Since the mapping from input to output in both $G$ and $D$ are not hard to quantify, both $G$ and $D$ are designed using the multilayer perceptron (MLP) architecture, with fully-connected hidden layers. In practice, the design of neural networks could also be further tailored depending on the specific situation. The parameters in both $G$ and $D$ will be updated according to the losses. In addition, $G(\mathbf{z})$ is considered as the sampled solution from the updating distribution $P_\mathbf{Z}(\mathbf{z})$. As the model converges, $P_\mathbf{Z}(\mathbf{z})$ will be similar as $P_{\text{data}}$. Instead of informing optimization parameters in $G$ and $D$, the optimization parameters are sent into the model-informed selector to ensure the feasibility and improve the optimality. In this way, the solutions generated from $G_m$, i.e., the solutions sampled from $P_{\text{data}}$, should be feasible and be close to the optimal solution.

### 3.2 Model-informed Selector

As mentioned in Sec. 3.1, the MI selector $M$ is proposed to select and update the feasible solutions among the generated solutions. Hence, the feasibility of generated solutions should be checked first. Afterwards, the filtered feasible solutions are further updated based on the objective function values and the gradient information, respectively. Accordingly, three different categories of new layers, including feasibility filter layer, comparison layer, and gradient-guided layer are proposed to establish $M$ in $G_m$. The demonstration of $M$ is shown in Figure 2.

In Figure 2, $M$ involves four steps to update the generated solutions. First, one feasibility filter layer is applied to check whether the solutions satisfy the given OPF constraints, i.e., feasibility check. Then a comparison layer compares the objective function values based on the saved solutions and feasible generated solutions. Afterwards, the gradient-guided layer updates the solutions according to the available gradient information. Since the feasibility of updated solutions may not be guaranteed, one more feasibility layer is applied to determine if the updated solutions are still feasible.

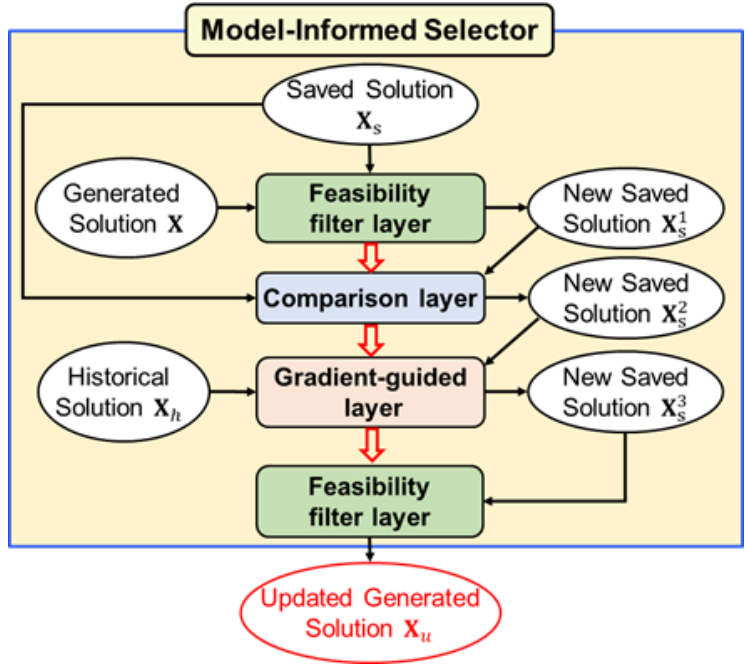

Figure 2 A demonstration of model-informed selector.

### 3.2.1 Feasibility filter layer

The feasibility filter layer is to check if the input solutions satisfy the constraints. Two identical feasibility filter layers are needed in $M$. Figure 3(a) provides a framework overview of the feasibility filter layer. Initially, the generated solutions $\mathbf{X}$ and saved solution $\mathbf{X}_s$ are sent to this layer as pairs. One solution in the pair is a generated solution, while the other is a saved solution. Afterwards, each pair will select one solution to be passed to the updated saved solution set $\mathbf{X}_s^1$.

In particular, as shown in Figure 3(a), the $i$-th pair of generated solution and saved solution, i.e., $(\mathbf{X}^{(i)}, \mathbf{X}_s^{(i)})$, may have four cases: i) feasible $\mathbf{X}^{(i)}$ and infeasible $\mathbf{X}_s^{(i)}$; ii) feasible $\mathbf{X}^{(i)}$ and feasible $\mathbf{X}_s^{(i)}$; iii) infeasible $\mathbf{X}^{(i)}$ and infeasible $\mathbf{X}_s^{(i)}$ also infeasible, as well as iv) infeasible $\mathbf{X}^{(i)}$ and feasible $\mathbf{X}_s^{(i)}$. For case i) and iv), the feasible solution will be passed to the updated saved solution set $\mathbf{X}_s^1$. However, in case ii) and iii), the feasibility of the generated solution and saved solution is the same. Since the generated solution is obtained from $G_m$ in current iteration, it is more representative than the saved solution to show the effectiveness of $G_m$. Hence, the generated solution should be passed to $\mathbf{X}_s^1$ in case ii) and iii). In this way, under the first three cases, the generated solutions from $\mathbf{X}$ will be passed to $\mathbf{X}_s^1$. Otherwise, the solutions from $\mathbf{X}_s$ will be passed to $\mathbf{X}_s^1$. Notably, the feasibility of each pair $(\mathbf{X}^{(i)}, \mathbf{X}_s^{(i)})$ is also recorded as a label, i.e., which feasibility case it belongs to, denoted as $S_i$ ($S_i = 1,2,3,4$) in a label sequence $S$, and it will be sent to the comparison layer.

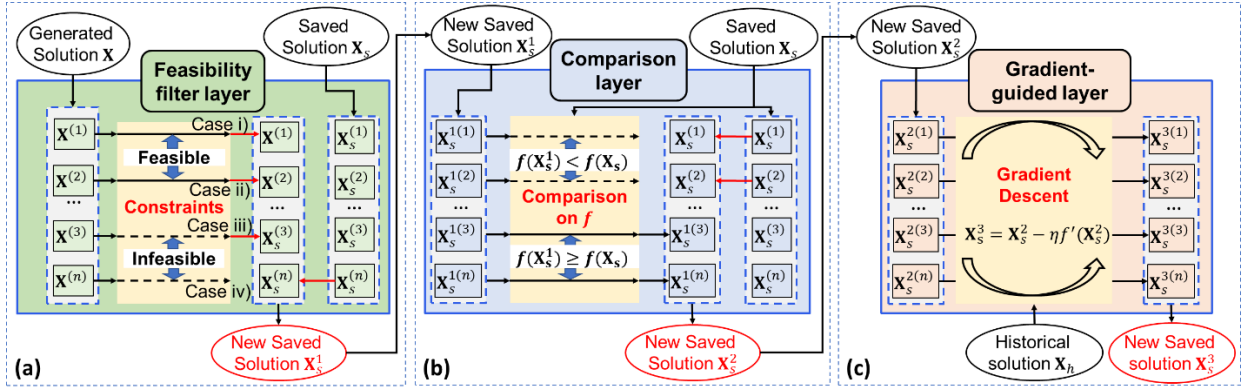

Figure 3 The framework of the feasibility filter layer (a), comparison layer (b), and gradient-guided layer (c) in the model-informed selector.

The algorithm for the feasibility filter layer can be demonstrated by Algorithm 1 below. For the $i$-th pair, only when $\mathbf{X}^{(i)}$ is infeasible and $\mathbf{X}_s^{(i)}$ is feasible, $\mathbf{X}_s^{(i)}$ is passed to $\mathbf{X}_s^1$ as $\mathbf{X}_s^{1(i)}$. Otherwise, $\mathbf{X}^{(i)}$ is assigned as $\mathbf{X}_s^{1(i)}$. Thus, in each iteration, all the solutions in $\mathbf{X}_s^1$ are essentially the generated solutions either synthesized from current iteration or previous iterations. In this way, the feasibility of the solutions which will be sent to $D$ could be guaranteed. The feasibility of $\mathbf{X}$ and $\mathbf{X}_s$ is also recorded as labels in $S$, which will be applied in the comparison layer.

**Algorithm 1:** Feasibility filter algorithm

**Input:** $\mathbf{X}\{\mathbf{X}^{(1)}, \mathbf{X}^{(2)}, \ldots, \mathbf{X}^{(n)}\}$, $\mathbf{X}_s\{\mathbf{X}_s^{(1)}, \mathbf{X}_s^{(2)}, \ldots, \mathbf{X}_s^{(n)}\}$, $\mathbf{A}$, $\mathbf{b}$
**Step 1:** Define a new saved solution set as $\mathbf{X}_s^1$
**Step 2:** Generate a $n$-element sequence $S: \{0,0,\ldots,0\}$
**For** $i = 1$ **to** $n$ **do**
  **Step 3:** Calculate $\mathbf{b} - \mathbf{AX}^{(i)}$
  If $\mathbf{b} - \mathbf{AX}^{(i)} \leq \mathbf{0}$ and $\mathbf{b} - \mathbf{AX}_s^{(i)} > \mathbf{0}$:
    **Step 4:** Assign $\mathbf{X}^{(i)}$ to $\mathbf{X}_s^{1(i)}$ and set $S_i = 1$
  Else if $\mathbf{b} - \mathbf{AX}^{(i)} \leq \mathbf{0}$ and $\mathbf{b} - \mathbf{AX}_s^{(i)} \leq \mathbf{0}$:
    **Step 5:** Assign $\mathbf{X}^{(i)}$ to $\mathbf{X}_s^{1(i)}$ and set $S_i = 2$
  Else if $\mathbf{b} - \mathbf{AX}^{(i)} > \mathbf{0}$ and $\mathbf{b} - \mathbf{AX}_s^{(i)} > \mathbf{0}$:
    **Step 6:** Assign $\mathbf{X}^{(i)}$ to $\mathbf{X}_s^{1(i)}$ and set $S_i = 3$
  Else: Assign $\mathbf{X}_s^{(i)}$ to $\mathbf{X}_s^{1(i)}$ and set $S_i = 4$
**Output** $\mathbf{X}_s^1$, $S$

### 3.2.2 Comparison layer

The comparison layer is to compare the objective function values between $\mathbf{X}_s^1$ and $\mathbf{X}_s$. Figure 3(b) depicts the framework of the comparison layer in $M$. Initially, $\mathbf{X}_s^1$ and $\mathbf{X}_s$ are also sent as pairs to the comparison layer. One solution in the pair is from $\mathbf{X}_s^1$ while the other is from $\mathbf{X}_s$. Similar to the feasibility filter layer, each pair will select one solution to be passed to the newly saved solution set $\mathbf{X}_s^2$.

Denote that the objective function is $f(\cdot)$. The $i$-th pair of generated solution and saved solution, $(\mathbf{X}_s^{1(i)}, \mathbf{X}_s^{(i)})$, may have two cases, as shown in Eq. (9),

$$\text{i) } f(\mathbf{X}_s^{1(i)}) \geq f(\mathbf{X}_s^{(i)}); \text{ ii) } f(\mathbf{X}_s^{1(i)}) < f(\mathbf{X}_s^{(i)}). \tag{9}$$

Under such circumstances, a natural idea is to pass the solution with smaller objective function value to $\mathbf{X}_s^2$. However, the feasibility of $\mathbf{X}$ and $\mathbf{X}_s$ should also be considered in the comparison layer. Hence, cases i) and ii) are discussed separately according to the feasibility of $\mathbf{X}$ and $\mathbf{X}_s$:

(a) For the case i), if $\mathbf{X}_s^{(i)}$ is feasible, i.e., $S_i$ is 1 or 2, $\mathbf{X}_s^{(i)}$ will be assigned as $\mathbf{X}_s^{2(i)}$; Otherwise, $\mathbf{X}_s^{1(i)}$ will be sent to $\mathbf{X}_s^2$ as $\mathbf{X}_s^{2(i)}$.

(b) For the case ii), only when $\mathbf{X}_s^{1(i)}$ is infeasible and $\mathbf{X}_s^{(i)}$ is feasible, i.e., $S_i$ is 1, $\mathbf{X}_s^{(i)}$ will be assigned as $\mathbf{X}_s^{2(i)}$, Otherwise, $\mathbf{X}_s^{1(i)}$ will be sent to $\mathbf{X}_s^2$ as $\mathbf{X}_s^{2(i)}$.

Based on the above two cases, the algorithm for comparison layer is demonstrated in Algorithm 2 below. Two cases are considered separately with the help of $f(\mathbf{X}_s^{1(i)})$, $f(\mathbf{X}_s^{(i)})$, and $S$. In this way, the feasible solutions with smaller objective function values could be passed to $\mathbf{X}_s^2$.

---

**Algorithm 2:** Objective function value algorithm

---

**Input:** $\mathbf{X}_s^1 \{\mathbf{X}_s^{1(1)}, \mathbf{X}_s^{1(2)}, \ldots, \mathbf{X}_s^{1(n)}\}$, $\mathbf{X}_s \{\mathbf{X}_s^{(1)}, \mathbf{X}_s^{(2)}, \ldots, \mathbf{X}_s^{(n)}\}$, $f(\cdot)$, $S$
**Step 1:** Define a new saved solution set as $\mathbf{X}_s^2$
**For** $i = 1$ **to** $n$ **do**
  **Step 2:** Calculate $f(\mathbf{X}_s^{1(i)})$ and $f(\mathbf{X}_s^{(i)})$
  If $f(\mathbf{X}_s^{1(i)}) < f(\mathbf{X}_s^{(i)})$:
    If $S_i = 1$: Assign $\mathbf{X}_s^{(i)}$ to $\mathbf{X}_s^{2(i)}$; Else: Assign $\mathbf{X}_s^{1(i)}$ to $\mathbf{X}_s^{2(i)}$
  Else if $S_i > 2$: Assign $\mathbf{X}_s^{1(i)}$ to $\mathbf{X}_s^{2(i)}$; Else: Assign $\mathbf{X}_s^{(i)}$ to $\mathbf{X}_s^{2(i)}$
**Output** $\mathbf{X}_s^2$

### 3.2.3 Gradient-guided layer

The gradient-guided layer is to update the generated solutions according to the available gradient information. Figure 3(c) depicts the framework of the gradient-guided layer in $M$. Initially, this layer is fed by $\mathbf{X}_s^2$ and historical solution $\mathbf{X}_h$, where both $\mathbf{X}_s^2$ and $\mathbf{X}_h$ have the same number of solutions. Hence, $\mathbf{X}_s^2$ and $\mathbf{X}_h$ are also combined as pairs in this layer. The $i$-th pair consists of the $i$-th solution of $\mathbf{X}_s^2$, $\mathbf{X}_s^{2(i)}$, and the $i$-th solution of $\mathbf{X}_h$, $\mathbf{X}_h^{(i)}$. Afterwards, $\mathbf{X}_s^{2(i)}$ will be updated to $\mathbf{X}_s^{3(i)}$ based on $\mathbf{X}_h^{(i)}$. Then $\mathbf{X}_s^{3(i)}$ will be sent to the updated saved solution set $\mathbf{X}_s^3$ as the $i$-th solution.

The solutions in $\mathbf{X}_s^2$ are updated using gradient descent (GD) algorithm (Ruder 2017). However, due to the complex formats of objective functions in OPF, $\nabla f$ may be hard to calculate. Hence, for $\mathbf{X}_s^{2(i)}$, $\nabla f$ is estimated by calculating the slope between $\left(\mathbf{X}_s^{2(i)}, f\left(\mathbf{X}_s^{2(i)}\right)\right)$ and $\left(\mathbf{X}_h^{(i)}, f\left(\mathbf{X}_h^{(i)}\right)\right)$ in this work. Afterwards, the step size of the GD algorithm is denoted as $\eta$, which could be determined through experimental trials. In this way, the solutions in $\mathbf{X}_s^3$ could be updated toward the direction with smaller objective function values. Notably, the gradient-guided layer is not mandatory in MI-GAN, but it could significantly accelerate the training process due to the guidance from gradient. Hence, MI-GAN will also work without this layer if the OPF model is highly complex and the gradient is hard to quantify.

---

**Algorithm 3:** Model-informed selector algorithm

---

**Input:**
$\mathbf{X}\{\mathbf{X}^{(1)}, \mathbf{X}^{(2)}, \dots, \mathbf{X}^{(n)}\}$, $\mathbf{X}_h\{\mathbf{X}_h^{(1)}, \mathbf{X}_h^{(2)}, \dots, \mathbf{X}_h^{(n)}\}$, $\mathbf{X}_s\{\mathbf{X}_s^{(1)}, \mathbf{X}_s^{(2)}, \dots, \mathbf{X}_s^{(n)}\}$, $f(\cdot)$, $\mathbf{A}$, $\mathbf{b}$, parameter $\eta$
**Step 1:** Obtain $\mathbf{X}_s^1$, $S$ by sending $\mathbf{X}$, $\mathbf{X}_s$, $\mathbf{A}$, $\mathbf{b}$ to **algorithm 1**
**Step 2:** Obtain $\mathbf{X}_s^2$ by sending $\mathbf{X}_s^1$, $\mathbf{X}_s$, $f(\cdot)$, $S$ to **algorithm 2**
**Step 3:** Replicate $\mathbf{X}_s^2$ as $\mathbf{X}_s^3$
**For** $i = 1$ **to** $n$ **do**
  **Step 4:** Calculate $f(\mathbf{X}_h^{(i)})$ and $f(\mathbf{X}_s^{2(i)})$
  If $\left(f(\mathbf{X}_s^{2(i)}) - f(\mathbf{X}_h^{(i)})\right) \leq 0$:
    **Step 5:** Obtain $\mathbf{X}_s^{3(i)} = \mathbf{X}_s^{2(i)} + \eta \ (f(\mathbf{X}_s^{2(i)})\text{-}f(\mathbf{X}_h^{(i)})) / (\mathbf{X}_s^{2(i)}\text{-}\mathbf{X}_h^{(i)})$
  Else:
    **Step 6:** Obtain $\mathbf{X}_s^{3(i)} = \mathbf{X}_s^{2(i)} - \eta \ (f(\mathbf{X}_s^{2(i)})\text{-}f(\mathbf{X}_h^{(i)})) / (\mathbf{X}_s^{2(i)}\text{-}\mathbf{X}_h^{(i)})$
**Step 7:** Obtain $\mathbf{X}_u$ by sending $\mathbf{X}_s^3$, $\mathbf{X}_s^2$, $\mathbf{A}$, $\mathbf{b}$ to **algorithm 1**
**Output** $\mathbf{X}_u$

---

In summary, based on the above-mentioned three new layers, the algorithm to implement $M$ in $G_m$ is demonstrated in Algorithm 3. The generated solution set $\mathbf{X}$ is updated as $\mathbf{X}_s^1$ in the first feasibility filter layer. Afterwards, $\mathbf{X}_s^1$ is updated as $\mathbf{X}_s^2$ in the comparison layer, and then $\mathbf{X}_s^2$ is updated as $\mathbf{X}_s^3$ in the gradient-guided layer. Finally, $\mathbf{X}_s^3$ is updated as $\mathbf{X}_u$ in the second feasibility filter layer. Notably, the input in the second feasibility filter layer is $\mathbf{X}_s^3$ and $\mathbf{X}_s^2$ rather than $\mathbf{X}$ and $\mathbf{X}_s$. In this way, $\mathbf{X}_u$ is considered as the output of model-informed selector. Thus, in each iteration, the generated solutions could be updated as feasible solutions towards the direction with smaller (i.e., better) objective function values.

### 3.3 Property of MI-GAN and Its Requirement

Based on the proposed MI-selector, the algorithm for MI-GAN is shown in Algorithm 4. The actual solution set, i.e., the historical solution set, $\mathbf{X}_h$, and the initialized saved solution set, i.e., the saved solution set $\mathbf{X}_s$, are sent to the MI-GAN. With the support of historical solutions, when $L_D$ and $L_{G_m}$ converge, the generated solutions should be similar to the historical solutions. In addition, the objective function values of the saved solutions decrease with each iteration. In this way, the generated solutions could gradually approach the optimal solution.

---

**Algorithm 4:** MI-GAN algorithm for OPF
---
**Input:** $\mathbf{X}_h$ $\{\mathbf{X}_h^{(1)}, \mathbf{X}_h^{(2)}, ..., \mathbf{X}_h^{(m)}\}$, $\mathbf{X}_s$ $\{\mathbf{X}_s^{(1)}, \mathbf{X}_s^{(2)}, ..., \mathbf{X}_s^{(n)}\}$, $f(\cdot)$, $\mathbf{A}$, $\mathbf{b}$, $f$, parameter $\eta$, $n$
**Repeat**
   **Step 1:** Randomly generate noise $\mathbf{Z}$
   **Step 2:** Generate $n$ fake solutions by generator as $\mathbf{X}$ $\{\mathbf{X}^{(1)}, \mathbf{X}^{(2)}, ..., \mathbf{X}^{(n)}\}$
   **Step 3:** Obtain $\mathbf{X}_u$ by inputting $\mathbf{X}_h$, $f(\cdot)$, $\mathbf{A}$, $\mathbf{b}$, $f$ to **algorithm 3**
   **Step 4:** Send $\mathbf{X}_h$ and $\mathbf{X}_u$ into discriminator $D$ to get output label results $D(\mathbf{X})$ and $D(\mathbf{X}_u)$, respectively
   **Step 5:** Optimize the model parameters based on the output of discriminator
   **Step 6:** Assign $\mathbf{X}_u$ to $\mathbf{X}_s$
**Until** both the $\text{Loss}_{G_m}$, $-D(\mathbf{X})$, and $\text{Loss}_D$, $D(\mathbf{X}) - D(\mathbf{X}_u)$, converge
   **Output** $G_m$, $D$

---

In the MI-GAN, the initialization requires to pre-define the saved solution set in the first iteration. Under such circumstances, an initialized saved solution set is considered, among which each solution is set as a vector with extremely large values. In this way, based on the feasibility layer and comparison layer, the initialized solutions in the initialized saved solution set could be replaced by the generated solutions in the first iteration. Then in the following iterations, all the solutions in $\mathbf{X}_s$ are the generated solutions from MI-GAN, either synthesized from current iteration or previous iterations. Hence, the initialized saved solutions will not interfere with the update of MI-GAN.

It is worth noting that MI-GAN can also work well even if the historical solutions are not available. If there are no historical solutions, with the given OPF problem, sampling from the feasible region could be applied to obtain the feasible solutions, which could be considered as actual solutions. However, in the large-scale OPF models, the sampling for feasible solutions may take a long time. Under such circumstances, in order to reduce the sampling time, the constraints could be partially relaxed. The case studies in Sec. 4 will show that the proposed method can still work well with the relaxed constraints.

Considering that the proposed MI-generator, $G_m$, is employed, it is also essential to prove that, as shown in Proposition 1, the convergence property is similar to GAN.

***Proposition 1.*** The convergence property of MI-GAN is similar to GAN, i.e., when $P_{\text{data}} = P_Z$ (Goodfellow et al. 2014), it will converge.

***Proof.*** Actual solutions are from the distribution of historical solutions, $P_{\text{data}}$. The artificial solutions are generated and selected from another distribution, $P_Z$. When $G_m$ is fixed, the optimal discriminator $D$ can be shown as

$$D_G^*(\mathbf{x}) = \frac{P_{\text{data}}(\mathbf{x})}{P_{\text{data}}(\mathbf{x}) + P_Z(\mathbf{x})} \tag{10}$$

Then the minimax game when $G_m$ is fixed is reformulated as

$$\max_D V(D, G_m) = \mathbb{E}_{\mathbf{x} \sim P_{\text{data}}(\mathbf{x})} \left[ \frac{P_{\text{data}}(\mathbf{x})}{P_{\text{data}}(\mathbf{x}) + P_Z(\mathbf{x})} \right] + \mathbb{E}_{\mathbf{x} \sim P_Z(\mathbf{x})} \left[ \frac{P_Z(\mathbf{x})}{P_{\text{data}}(\mathbf{x}) + P_Z(\mathbf{x})} \right] \tag{11}$$

Afterwards, only when $P_{\text{data}} = P_Z$, the global minimum value of $\max_D V(D, G_m)$ shown in Eq. (11) can be achieved (Goodfellow et al. 2014). In this way, the distribution of the generated solutions will gradually approach the distribution of the historical solutions. *Q.E.D.*

Based on the convergence property, $P_Z$ should be similar to $P_{\text{data}}$. When model converges, the solutions could be sampled from the model-informed generator and the best solution could be chosen. Denote that the optimal solution is $\mathbf{x}^*$. Hence, if $\mathbf{x}^*$ locates in $P_{\text{data}}$, the generated solutions from $G_m$, i.e., $\mathbf{X}$, should be around $\mathbf{x}^*$. Following that, the solution with the smallest objective function values among $\mathbf{X}$ could be selected as the output solution from the MI-GAN.

However, if $\mathbf{x}^*$ does not locate in $P_{\text{data}}$, $\mathbf{X}$ may not be around $\mathbf{x}^*$. Hence, a new algorithm to improve the training of MI-GAN called recursive iteration algorithm, is being developed in Sec. 3.4 to address this issue.

### 3.4 Recursive Iteration Algorithm

As described in Sec. 3.3, when the historical solutions $\mathbf{X}_h$ are not available, the sampled actual feasible solutions applied to train the MI-GAN may be far away from $\mathbf{x}^*$. Denote such actual solution set as $\mathbf{X}_a$. Though the MI-GAN converges, the generated solutions $\mathbf{X}$ probably are still not close enough to $\mathbf{x}^*$. Under such circumstances, it is important to make sure that the generated solutions $\mathbf{X}$ will approach to $\mathbf{x}^*$. Hence, as shown in Figure 4, a recursive iteration algorithm for MI-GAN update is proposed to gradually bring $\mathbf{X}$ closer to $\mathbf{x}^*$. Since $\mathbf{X}$ are similar to $\mathbf{X}_a$, a natural idea is to select the solutions with smallest objective function values from both $\mathbf{X}$ and $\mathbf{X}_a$. Such updated solutions, namely, $\mathbf{X}_a'$, should be closer to $\mathbf{x}^*$ than $\mathbf{X}_a$.

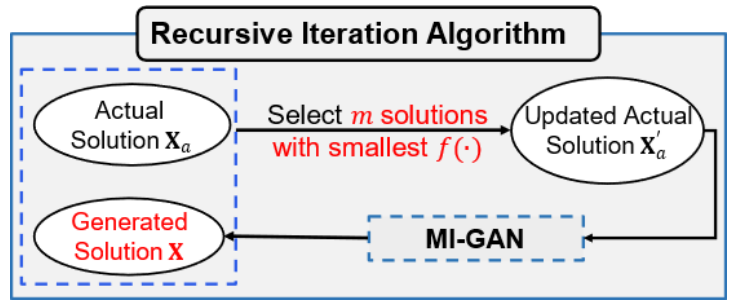

Figure 4 The framework of the recursive iteration algorithm

Afterwards, $\mathbf{X}_a'$ continue to train MI-GAN until the losses converge again. Each loop, which could be considered as one recursive iteration, will make newly generated solutions closer to $\mathbf{x}^*$ than the previously generated solutions. It is also important to note that, only the feasible generated solutions $\mathbf{X}$ are selected to next recursive iteration after they pass the feasibility filter layer in the model-informed selector. Specifically, the feasibility filter layer is directly informed by the OPF constraints so that the feasibility of $\mathbf{X}$ is not influenced by $\mathbf{X}_a$. Therefore, $\mathbf{X}_a'$ selected from $\mathbf{X}$ and $\mathbf{X}_a$ must be a better actual solution set than $\mathbf{X}_a$ since it is both feasible and having smaller objective function values. To further demonstrate the effectiveness of the recursive iteration, Proposition 2 is presented below.

***Proposition 2.*** Denote that the output solution from the $k$-th trained MI-GAN with the smallest $f(x)$ is $\hat{\mathbf{x}}_k$. Then $\hat{\mathbf{x}}_k$ will not be worse than $\hat{\mathbf{x}}_{k-1}$. That is,

$$f(\hat{\mathbf{x}}_k) \leq f(\hat{\mathbf{x}}_{k-1}) \tag{12}$$

***Proof.*** Denote that the actual solution set and generated solution set of the $k$-th trained MI-GAN is $\mathbf{X}_{a_k}$ and $\mathbf{X}_k$. According to Figure 4, $\mathbf{X}_{a_k}$ are selected among $\mathbf{X}_{a_{k-1}}$ and $\mathbf{X}_{k-1}$. Since the selection principle is to compare the objective function values, $\mathbf{X}_{a_k}$ is better than $\mathbf{X}_{a_{k-1}}$. Therefore, for any solution $\mathbf{x}_k$ from $\mathbf{X}_{a_k}$, there must exist at least one solution $\mathbf{x}_{k-1}$ from $\mathbf{X}_{a_{k-1}}$ such that $f(\mathbf{x}_k) \leq f(\mathbf{x}_{k-1})$. *Q.E.D.*

Based on proposition 2, more importantly, to show the convergence property of the proposed recursive iteration algorithm for MI-GAN, proposition 3 is obtained below as well.

***Proposition 3.*** $\hat{\mathbf{x}}_k$ will gradually converge to optimal solution $\mathbf{x}^*$ as $k$ increases.

***Proof.*** Due to proposition 2, $f(\hat{\mathbf{x}}_k)$ monotonically decreases as $k$ increases. In addition, $f(\hat{\mathbf{x}}_k)$ is bounded by $f(\mathbf{x}^*)$. According to the monotone convergence theorem (Kolmogorov 1957), $\hat{\mathbf{x}}_k$ could converge to $\mathbf{x}^*$ as $k$ increases. *Q.E.D.*

In practice, since the optimal solution is unknown, it is critical to define the stopping criteria for the recursive iteration algorithm. Due to the gradient-guided layer, the output solutions of MI-GAN will typically outperform $\mathbf{X}_a$. Then Eq. (13) could be satisfied in most recursive iterations, which will make $\hat{\mathbf{x}}_k$ gradually approach to $\mathbf{x}^*$.

However, when $\hat{\mathbf{x}}_k$ is around $\mathbf{x}^*$, then Eq. (13) may not be satisfied. Hence, based on Eq. (13), the stopping criteria could be developed as follows: if $f(\hat{\mathbf{x}}_k)$ does not decrease in next two consecutive iterations, i.e., no better solutions, the iterative process will be terminated.

$$|\mathbf{x}^* - \hat{\mathbf{x}}_k| \leq |\mathbf{x}^* - \hat{\mathbf{x}}_{k-1}| \tag{13}$$

According to Figure 4 and the stopping criteria, the pseudo code for the recursive iteration algorithm is demonstrated in Algorithm 5 below. In each iteration, the MI-GAN is trained according to $\mathbf{X}_a$ and generate $\mathbf{X}$ after the model converges. Then $m$ solutions are selected as $\mathbf{X}'_a$ among $\mathbf{X}_a$ and $\mathbf{X}$. $\mathbf{X}'_a$ is applied in the next recursive iteration, and iteration will be terminated based on the proposed stopping criteria. In this way, this recursive iteration algorithm could help to find a solution close to optimal.

---

**Algorithm 5:** Recursive iteration algorithm

---

**Input:** $\mathbf{X}_a \{\mathbf{X}_a^{(1)}, \mathbf{X}_a^{(2)}, \dots, \mathbf{X}_a^{(m)}\}$, parameter $\delta$, $h$
**Step 1:** Set $k = 1$
**Loop**
   **Step 2:** Train MI-GAN by $\mathbf{X}_a$
   **Step 3:** Generate $h$ artificial samples $\mathbf{X}$ from the trained MI-GAN
   **Step 4:** Select $m$ solutions with smallest objective function value as $\mathbf{X}'_a$ among $\mathbf{X}_a$ and $\mathbf{X}$
   **Step 5:** Assign $\mathbf{X}'_a$ to $\mathbf{X}_a$
   **Step 6:** Calculate the minimum objective function value of $\mathbf{X}_a$, $f_{min}^k$ and its output solution $\hat{\mathbf{x}}_k$
   **Step 7:** $k = k + 1$
**Until** $f_{min}^{k-2} \leq f_{min}^{k-1}$ **and** $f_{min}^{k-2} \leq f_{min}^k$:
   **Output** $f_{min}^{k-2}$ and $\hat{\mathbf{x}}_{k-2}$

---

## 4 Case Studies

### 4.1 Data Introduction and Experimental Setup

The experimental setups in this work are based on (Venzke et al. 2020), and six DC-OPF cases are demonstrated. The detailed setup of each case is shown in Table 1 (Zimmerman, Murillo-Sánchez, and Thomas 2010). As the number of buses, i.e., $n_b$, increases from 9 to 162, so does the number of constraints, from 24 to 592. Due to the significantly different scale of the cases, it could demonstrate the effectiveness of the proposed method comprehensively.

*Table 1: Experiment Setup for DC-OPF cases*

| Setup | Cases | | | | | |
|---|---|---|---|---|---|---|
| | Case9 | Case30 | Case39 | Case57 | Case118 | Case162 |
| $n_b$ | 9 | 30 | 39 | 57 | 118 | 162 |
| $n_g$ | 3 | 2 | 10 | 4 | 19 | 12 |
| $n_d$ | 3 | 21 | 21 | 42 | 99 | 113 |
| $n_{line}$ | 9 | 41 | 46 | 80 | 186 | 284 |
| Max. loading | 315.0MW | 283.4MW | 6254.2MW | 1250.8MW | 4242.0MW | 7239.1MW |
| Number of constraints | 24 | 86 | 112 | 168 | 410 | 592 |
| Number of relaxed constraints | 0 | 0 | 6 | 1 | 19 | 49 |

In order to obtain the sampled actual solutions for the proposed method as the input, 3,000 feasible solutions are initially sampled for each case before training the proposed method. In the sampling process, $\mathbf{p}_g$ is first uniformly sampled from the range defined in Eq. (5). Afterwards, $\boldsymbol{\theta}$ is calculated based on Eq. (3) and sent to Eq. (4) for feasibility check. Such sampled feasible solutions may elongate the training time due to the large number of recursive iterations. However, in the real-world applications of OPF problems, the historical solutions for the optimization models from previous operating days are usually available, which could be applied for optimization model of the current operating day. Specifically, such historical solutions could usually be considered as the solutions sampled from a distribution that is in proximity to the optimal solution unless there are big changes for the operating decisions. Therefore, the training time could be significantly reduced in the real-world applications.

As $n_b$ increases in each case, the difficulties of sampling the feasible solutions also increase, resulting in a significant increase of sampling time. Besides, it will also be more difficult for the proposed method to learn the data distribution and synthesize feasible solutions in the initial training stage. Hence, to reduce the sampling time and the difficulties to obtain sufficient feasible solutions, the constraints could be randomly relaxed as the number of variables is relatively large. The number of relaxed constraints for each case is also shown in Table 1, which is determined by experimental trials. The smallest number of relaxed constraints that ensures the proposed method MI-GAN to learn the actual data distribution is selected. It is also worth noting that, though some constraints are relaxed during the training process, the selected solution from the model are still needed to satisfy all the constraints after the model converges.

Notably, the proposed MI-GAN is compatible with most of the popular GAN architectures. In this study, MI-GAN is built by incorporating the popular Wasserstein GAN (WGAN) (Arjovsky, Chintala, and Bottou 2017) since WGAN could make the training process more stable than the conventional GAN. Besides, the multilayer perceptron (MLP) network is applied in both generator and discriminator of the proposed MI-GAN. All above cases have the same parameter setups when using the proposed MI-GAN. The generator consists of five fully connected layer, whereas the discriminator consists of three fully connected layer. The last layer in the generator does not use the activation function while all the other layers incorporate Leaky_ReLU as the activation function. TensorFlow 2.12.0 (Abadi et al. 2016) is applied for the training of the proposed method MI-GAN. The batching size is set as 50 while the number of iterations is set as 2,000. Each case involves five trials. In each trial, the output is one feasible solution with the smallest objective function value. Afterwards, the mean of the output solutions from different trials are calculated for discussion. The experiments were conducted on the Google colab with Python 3.10.12 (Bisong 2019).

According to Eq. (3) in Sec. 2, if $\mathbf{p}_g$ is obtained, $\boldsymbol{\theta}$ could also be calculated. Hence, to simplify the generation process of feasible solutions, $\mathbf{p}_g$ is generated in the generator of the proposed method instead of $\{\mathbf{p}_g, \boldsymbol{\theta}\}$. Then prior to passing the variables to the layers in the model-informed selector, $\boldsymbol{\theta}$ is initially calculated. In this way, the discriminator in the proposed MI-GAN will still distinguish the entire solution.

In this study, to demonstrate the effectiveness of the proposed method MI-GAN comprehensively, two different scenarios are considered. Sec. 4.2 showed the performance of MI-GAN for a given fixed OPF problem while Sec. 4.3 showed the performance of MI-GAN for the OPF problem with uncertainties, i.e., OPF problem with dynamic net loads.

## 4.2 MI-GAN performance for given fixed OPF problems

### 4.2.1 MI-GAN performance under small-scale DC-OPF problems

The performance of the proposed MI-GAN for the six given cases where $\rho = 1$ is shown in Table 2. In this study, under each case, the feasible solution generated from the MI-GAN with the smallest objective function value is considered as its output solution. The accuracy of the output solution is measured by mean absolute errors (MAE) in percentage from the objective function values of the output solution and the actual optimal solution. The MAEs for different cases are shown in the first row of Table 2, with the values in the brackets representing the standard deviations. Based on the MAEs, it is shown that the bias between the optimal objective function values from MI-GAN and the actual optimal objective function value could be less than 5% in most cases. Specifically, under the case30 and case57, the bias of the objective function values from the proposed method could be less than 1%. For case162, it is worth noting that the Euclidean distance between the output solution from MI-GAN and the actual optimal solution does not significantly increase, compared with the Euclidean distance calculated under other cases. Hence, though the MAE calculated under case162 is greater than 5%, it is still comparable. Therefore, the proposed method's output solutions are very close to the actual optimal solutions, demonstrating the method's effectiveness. Besides, the standard deviations under each case are less than 3%. Hence, it shows that MI-GAN could generate desired solutions accurately and consistently.

*Table 2: Performance of the proposed method for DC-OPF cases*

| Results | Cases | | | | | |
|---|---|---|---|---|---|---|
| | Case9 | Case30 | Case39 | Case57 | Case118 | Case162 |
| MAE (%) | 4.5 (2.9) | 0.06 (0.3) | 4.7 (1.6) | 0.8 (0.4) | 4.0 (2.6) | 9.4 (2.0) |
| Number of recursive iterations | 2.8 (1.6) | 1.2 (0.4) | 1.4 (0.5) | 2.4 (0.5) | 1.6 (0.8) | 3.8 (1.9) |
| Running time (sec) | 28.30 (1.91) | 28.99 (1.58) | 29.32 (0.17) | 32.39 (1.24) | 40.91 (0.80) | 43.26 (0.97) |

In addition, to demonstrate the effectiveness of the proposed recursive iteration algorithm, the number of recursive iterations is also shown in Table 2, where the values in the brackets are also the standard deviations. Except for case162, all cases could obtain near-optimal solutions within three recursive iterations. Though the number of recursive iterations in case162 is slightly larger, it is still less than four, which is comparable. Hence, it shows that the recursive iteration algorithm is beneficial and essential to improve MI-GAN.

Besides, the experimental running time of MI-GAN was also recorded to demonstrate the efficiency for the proposed method for use in the online power systems. Specifically, the running time for one recursive iteration under each case is collected from five replicates. Then the average running time is calculated. The running time and its standard deviations of the proposed method are also shown in Table 2. The running time will gradually increase as the number of variables increases, owing to the matrix calculations in the model-informed selector becoming increasingly complex.

Additionally, we conducted a comparative analysis by calculating the running time of the classical simplex algorithm (Dantzig 1990) when applied to solve identical OPF models. This comparison was essential to highlight the superior time efficiency of MI-GAN. To ensure a fair comparison, the simplex method is implemented by the linprog function in Python. The comparisons are demonstrated in Figure 5(a)-(b). Figure 5(a) depicts the average running time and the fitted curve of running time for both approaches. The running time of simplex method is exponential growth since the worst case needs to inspect all the vertex. Hence, it is fitted by an exponential regression model. As for the proposed method, the neural network does not significantly change for different cases and the increment of the running time is mostly due to the dimension increment in matrix calculation. Then the running time of the proposed MI-GAN is fitted by a linear regression model. It is well known that simplex algorithm for linear programming has an exponential worst case time complexity as it requires inspection of all vertices in the worst-case scenario (Klee and Minty 1972). We can also observe from Figure 5(a)-(b) that the running time of the simplex method exhibits exponential growth. As for the proposed MI-GAN, the neural network architecture does not significantly change for different cases and the increment of the running time is primarily attributed to the expansion in matrix calculations due to varying problem dimensions. Then the running time of the proposed MI-GAN was appropriately modeled using a linear regression model. As shown in Figure 5(a), since $R^2$ of the fitted running time curve for both approaches are greater than 90%, the equations to demonstrate the running time for both approaches are convincing. Hence, the running time of the proposed MI-GAN increases linearly while the running time of the simplex method increases exponentially. For the current cases, due to the essential matrix calculations within the neural networks, the proposed MI-GAN requires more time to execute than the simplex method. However, when analyzing the fitted running time curve, it becomes evident that the extended runtime of the MI-GAN is primarily influenced by certain persistent calculations that do not significantly increase with the number of variables. Therefore, as the number of variables increases to a larger scale, the running time of the proposed method could be much less than the simplex method, which is also proved in Sec. 4.2.2. In addition, the slopes of running time for both approaches, as shown in Figure 5(b), can also demonstrate the increasing trend of running time. The slope of the proposed method fluctuates whereas the slope of the simplex method increases. In case162, the slope of the simplex method is almost the same as the slope of the proposed method MI-GAN. Hence, MI-GAN may have a significantly shorter running time than the simplex method for large-scale cases. Since the proposed method could still obtain near optimal solutions, it has the full potential to be applied in large-scale power systems.

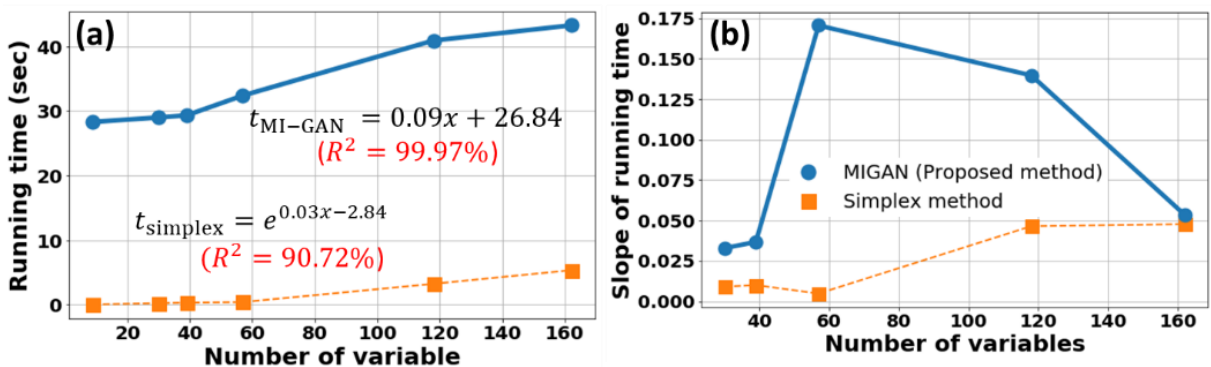

Figure 5 Comparison of MI-GAN and simplex method for running time (a) and the slope of running time (b).

4.2.2  MI-GAN performance under large-scale and nonlinear OPF problems

As described in Sec. 4.2.1, the proposed method MI-GAN should have better computational performance under large-scale power systems and be able to handle the nonlinear OPF problems as well. Therefore, to further demonstrate the performance of the proposed method from these two aspects, two more cases were conducted in this subsection accordingly.

The DC-OPF case2736 from Polish system during summer 2004 peak conditions (Zimmerman, Murillo-Sánchez, and Thomas 2010) is applied to demonstrate the superior time efficiency of the proposed MI-GAN method under large-scale OPF problems. The detailed setup of case2736 is shown in Table 3. The number of buses has increased to 2736 while the number of generators has increased to 420, which is much larger than the other cases in Sec. 4.2.1. As described in Sec. 4.1, to reduce the sampling time and the difficulties to obtain sufficient feasible solutions, 1000 constraints are relaxed in this case. Regarding the time efficiency, the proposed method takes 265.89 secs (with standard deviations 2.11 secs), which well fits the linear model of running time in Figure 5(a). As for the simplex method, 50,000 secs are still unable to obtain the optimal solution. Such long time also shows that the running time of the simplex should be increasing exponentially. Therefore, the running time of the proposed method is much smaller than the simplex for this case, which fully demonstrates that MI-GAN has great potential to advance the computational efficiency of solving OPF problems.

*Table 3: Experiment Setup for DC-OPF case2736 and AC-OPF case9*

| Setup | Cases | |
|---|---|---|
| | DC-OPF Case2736 | AC-OPF Case9 |
| $n_b$ | 2736 | 9 |
| $n_g$ | 420 | 3 |
| $n_d$ | 2011 | 3 |
| $n_{line}$ | 3504 | 9 |
| Number of constraints | 7848 | 72 |
| Number of relaxed constraints | 1000 | 0 |

In addition, to validate the performance of the proposed method MI-GAN under nonlinear OPF problems, an alternative current OPF (AC-OPF) instance of case9 is also demonstrated (Zimmerman, Murillo-Sánchez, and Thomas 2010). The detailed setup is shown in Table 3. Specifically, compared with DC-OPF problems, two new groups of variables, including the reactive power generation ($\mathbf{q}_g$) and the voltage magnitude ($\mathbf{v}$), are added to the AC-OPF problem, together with resulting nonlinear constraints. Therefore, there are 24 variables and 72 constraints, which is much larger than the DC-OPF problem of case9 though there are still only 9 buses. Besides, it is important to note that, since the gradient is hard to capture in the AC-OPF problems, the gradient-guided layer is not applied in this case. Therefore, the model-informed selector in the proposed MI-GAN consists of one feasibility filter layer and one comparison layer. In this case, though the number of variables and constraints has increased, the running time of the proposed method is still similar to the running time under DC-OPF of case9, i.e., about 30.65 secs (with stand deviations 0.25 secs). Hence, it shows that the nonlinear constraints do not significantly influence the time efficiency. Specifically, the output solution from the proposed method is feasible and near the optimal solution since the average MAE between the objective function value calculated by the output solution from MI-GAN and by the optimal solution is about 6.7% (with standard deviation 2.9%). Based on the relatively small MAE, it also demonstrates the potential effectiveness of the proposed method MI-GAN to handle AC-OPF.

### 4.3 Performance for OPF problem with uncertainties

#### 4.3.1 Solution accuracy

Given that the renewable energy intermittency may disturb power system stability, it is also important to demonstrate the robustness of the proposed method to handle OPF with uncertainties. Specifically, in our experiments, such uncertainties are considered as the dynamic changes of $\rho$, trying to investigate whether then proposed method MI-GAN could still have a good performance as $\rho$ changes. The experiments are conducted based on case9 and case57 and are designed as follows: the proposed method is initially trained given the original net loads, $\mathbf{p}_d$. After the proposed method converges, the net loads are changed to $\rho\mathbf{p}_d$. Under the net loads of $\rho\mathbf{p}_d$, the feasible region will change and thus the optimal solution will be $\mathbf{x}^{*\prime}$, resulting in the new objective function value, i.e., $f(\mathbf{x}^{*\prime})$. To identify this new optimal solution, the existing trained MI-GAN undergoes an updating process to adapt to the altered net loads. Following this, the MI-GAN should yield a new output solution, which should be close to $f(\mathbf{x}^{*\prime})$, within a single recursive iteration.

To better demonstrate the performance of the proposed MI-GAN, a benchmark approach based on the DNN regression (Falconer and Mones 2022) and conventional GAN (Goodfellow et al. 2014), which has been applied in several existing related studies, is selected to compare with the proposed MI-GAN. To make the comparison convincing, the network structure of the benchmark method is optimized as well. Specifically, in the benchmark method, MLP is also applied in both $D$ and $G$. Through hyper-parameter tuning, $G$ involves two fully-connected hidden layers while $D$ involves only one fully-connected hidden layer.

Since the feasible regions for both the proposed method MI-GAN and the benchmark method may change as the net loads change, the applied actual data may be neither feasible nor close to the optimal solution. Hence, when the net loads change, 1000 solutions that adhere to the updated constraints are sampled. Then, they will be added to the existing actual data to train the model. To fully show the capability of MI-GAN, two cases to update $\rho$ are considered, which are listed as follows:

i) Increasing the net loads $\mathbf{p}_d$ and $\rho$ is selected among the set $\{1.05, 1.1, 1.15, 1.2, 1.25, 1.3, 1.4, 1.5\}$.

ii) Decreasing the net loads $\mathbf{p}_d$ and $\rho$ is selected among the set $\{0.5, 0.6, 0.7, 0.75, 0.8, 0.85, 0.9, 0.95\}$.

Notably, research has extensively studied uncertainties in the day-ahead level using stochastic or robust optimization models. However, when it comes to the real-time level, resolving real-time OPF models has been the common practice when net loads increase or decrease. Our proposed MI-GAN provides a novel solution by directly incorporating these dynamic changes in net loads. While the initial training of MI-GAN is based on specific net loads, it possesses the ability to gradually update and adapt to find solutions for OPF problems with varying net loads. The MAEs and standard deviations of the proposed method and the benchmark approach for case9 are shown in Figure 6(a)-(d). The benchmark approach may have lower MAEs than the proposed method at the beginning when both increasing and decreasing the net loads. However, the MAEs of the benchmark approach will gradually increase and become larger than those of the proposed method as $|1-\rho|$ increases. Hence, it shows that the proposed method is more robust than the benchmark approach. In addition, the benchmark approach's standard deviations are generally higher than those of the proposed method and increase as $|1-\rho|$ increases, while the standard deviations of the proposed MI-GAN remain similar. The comparisons among standard deviations also reveal that the proposed method MI-GAN has higher stability than the benchmark approach.

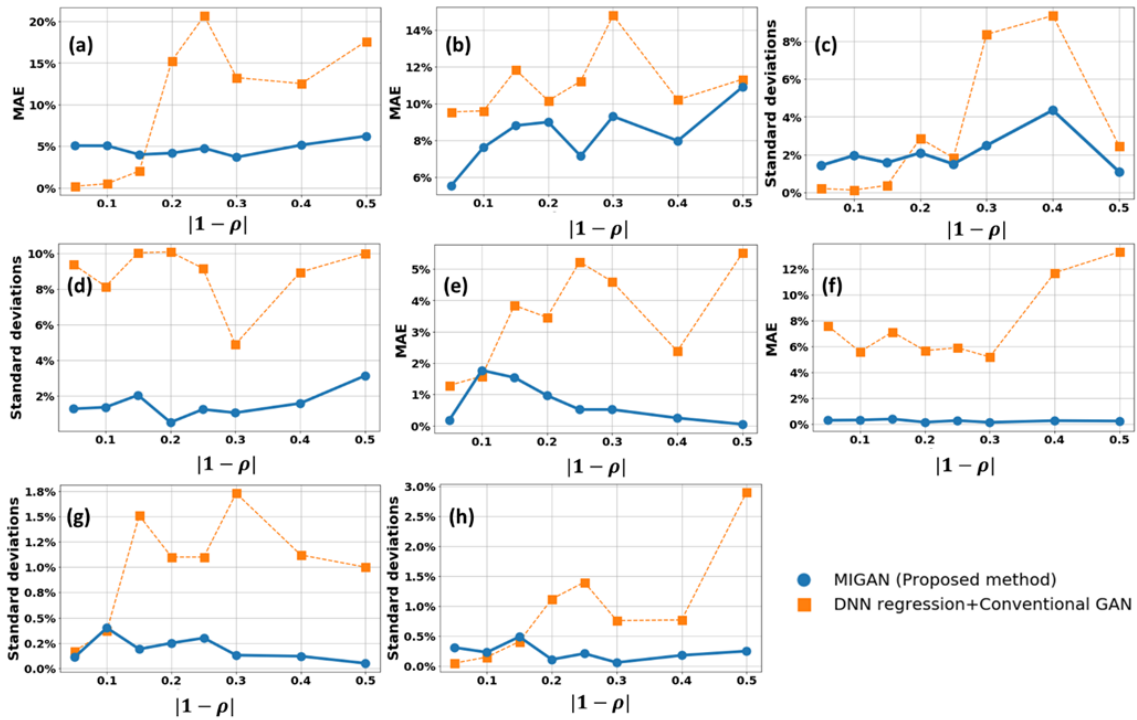

Figure 6 The comparison between the proposed method and DNN regression + Conventional GAN when increasing and decreasing the net loads for case9 (a-d) and case57 (e-h): (a) MAE when increasing the net loads under case9; (b) MAE when decreasing the net loads under case9; (c) standard deviations of MAE when increasing the net loads under case 9; (d) standard deviations of MAE when decreasing the net loads under case 9; (e) MAE when increasing the net loads under case57; (f) MAE when decreasing the net loads under case57; (g) standard deviations of MAE when increasing the net loads under case57; (d) standard deviations of MAE when decreasing the net loads under case57.

On the other hand, the MAEs and standard deviations of the proposed method and the benchmark approach for case57 are shown in Figure 6(e)-(h). It clearly shows that the MAEs of the benchmark approach will gradually increase and be larger than the proposed method regardless of the changing patterns of net loads. Furthermore, the MAEs and the standard deviations of the proposed method do not show any significantly increasing or decreasing as $|1-\rho|$ increases, i.e., as the dynamic changes become more dramatic, which also outperforms the benchmark method under case57. Overall, the experiments conducted under both case9 and case57 demonstrate that the proposed method is not sensitive to the dynamic changes of net loads. If a power system has uncertainties in dynamic changes of net loads (real-time level), the trained MI-GAN could still be applied and find the solutions effectively.

### 4.3.2 Needed recursive iterations

Since the proposed method employs the recursive iteration algorithm, it is also necessary to discuss the changes of recursive iterations as the power systems interfere. The number of recursive iterations for case9 and case57 are shown in Figure 7(a)-(b). For both case9 and case57, the number of recursive iterations is mostly around 2 to 5, and it does not change significantly as $|1-\rho|$ increases. Hence, it shows the high robustness of the proposed method. Furthermore, despite the fact that the benchmark approach is optimized and has less layers than the proposed method, the benchmark approach runs in about 55 seconds in both cases 9 and 57. Compared with the running time of the proposed method in Table 2, the proposed method still outperforms the benchmark approach in terms of efficiency. Therefore, the experiments with different net loads fully demonstrate the superior performance of the proposed method and its high potential to be applied in the large-scale OPF problems under uncertainty.

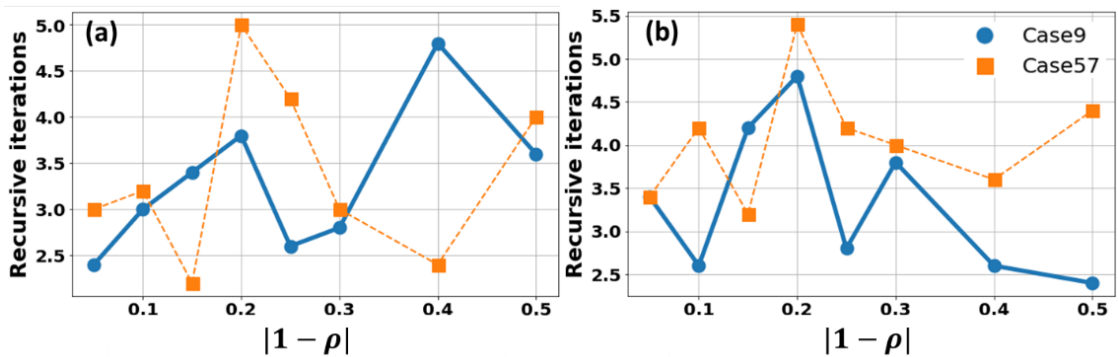

Figure 7 The number of recursive iterations when changing the power demands under case9 and case57: (a) the number of recursive iterations when increasing the power demands; (b) the number of recursive iterations when decreasing the power demands.

# 5  Conclusions

In this paper, a novel MI-GAN framework is proposed to address the OPF problems with uncertainty, specifically, the OPF problems with dynamic net loads, from a data-driven perspective. In comparison to the existing optimization models, the proposed MI-GAN has three major contributions: (i) three important new layers, including the feasibility filter layer, comparison layer, and gradient-guided layer, are proposed and designed to ensure the feasibility and improve the optimality of generated solutions; (ii) an efficient MI selector in conjunction with the three new layers are developed, incorporating the GAN-based architecture as an important component of the generator, i.e., MI-generator; and (iii) a new recursive iteration algorithm is also proposed to further reduce the bias between selected solutions and optimal solutions. It is also worth noting that the proposed MI-GAN framework is compatible with most of the popular GAN architectures, which could further improve its capability of solving different optimization problems in power systems.

The superior performance of MI-GAN is demonstrated by six DC-OPF cases, including case9, case30, case37, case57, case118, case162. Among these six cases, the MAEs between the objective function values obtained from the proposed method and the actual optimal objective function value demonstrate the effectiveness of the proposed MI-GAN. In addition, the running times of the proposed method are also discussed to show its potential of computational efficiency. The running time increases linearly as the number of variables increases, showing that the proposed MI-GAN is more efficient than the conventional optimization techniques in solving OPF problems with uncertainties. Furthermore, the proposed method also performed well under a large-scale DC-OPF instance and an AC-OPF instance, which show its high computational efficiency for large-scale power systems and strong capability for nonlinear optimization model. Besides, to demonstrate its capability of handling uncertainties, experiments with increasing and decreasing net loads are also conducted under case9 and case57. The lower MAEs of the proposed method than the benchmark approach demonstrate that the proposed method is effective to handle the power system dynamics. The relatively low standard deviations and relatively stable number of recursive iterations also show the high robustness of the proposed method. Therefore, the proposed method MI-GAN is very promising for finding the optimal solutions of the complex OPF-related problems with uncertainty.